\documentclass[10pt,twocolumn,letterpaper]{article}

\usepackage{cvpr}
\usepackage{times}
\usepackage{epsfig}
\usepackage{graphicx}
\usepackage{amsmath}
\usepackage{amssymb}
\usepackage{algorithm} 
\usepackage{algorithmic}
\usepackage{helvet}
\usepackage{courier}
\usepackage[english]{babel}
\usepackage{courier}
\usepackage{bm}
\usepackage{CJK}
\usepackage{indentfirst}
\usepackage{multirow}
\usepackage{blindtext}
\usepackage{color}
\usepackage{epsfig}

\usepackage[pagebackref=true,breaklinks=true,letterpaper=true,colorlinks,bookmarks=false]{hyperref}
\usepackage{breakurl}
\makeatletter

\def\cvprPaperID{****} 

\cvprfinalcopy
\begin{document}

\title{Deep Spatial Feature Reconstruction for Partial Person Re-identification: Alignment-free Approach}

\author{Lingxiao He$^{* 1,2}$, Jian Liang$^{* 1,2}$, Haiqing Li$^{1,2}$, and Zhenan Sun$^{1,2,3}$\\
$^{1}$ CRIPAC \& NLPR, CASIA $^{2}$ University of Chinese Academy of Sciences, Beijing, P.R. China\\
$^{3}$ Center for Excellence in Brain Science and Intelligence Technology, CAS\\
{\tt\small \{lingxiao.he, jian.liang, hqli, znsun\}@nlpr.ia.ac.cn}
}

\maketitle

\begin{abstract}
\let\thefootnote\relax\footnotetext{$^*$ Authors contributed equally.}
	Partial person re-identification (re-id) is a challenging problem, where only several partial observations (images) of people are available for matching.
    However, few studies have provided flexible solutions to identifying a person in an image containing arbitrary part of the body.
    In this paper, we propose a fast and accurate matching method to address this problem. The proposed method leverages Fully Convolutional Network (FCN) to generate fix-sized spatial feature maps such that pixel-level features are consistent. To match a pair of person images of different sizes, a novel method called Deep Spatial feature Reconstruction (DSR) is further developed to avoid explicit alignment. Specifically, DSR exploits the reconstructing error from popular dictionary learning models to calculate the similarity between different spatial feature maps. In that way, we expect that the proposed FCN can decrease the similarity of coupled images from different persons and increase that from the same person. Experimental results on two partial person datasets demonstrate the efficiency and effectiveness of the proposed method in comparison with several state-of-the-art partial person re-id approaches. Additionally, DSR achieves competitive results on a benchmark person dataset Market1501 with 83.58\% Rank-1 accuracy. The website of DSR code can be found from \url{https://github.com/lingxiao-he/Partial-Person-ReID}.

\end{abstract}
\section{Introduction}
\begin{figure}[t]
    \centering
       \vspace{0em}
    \includegraphics[width=8cm]{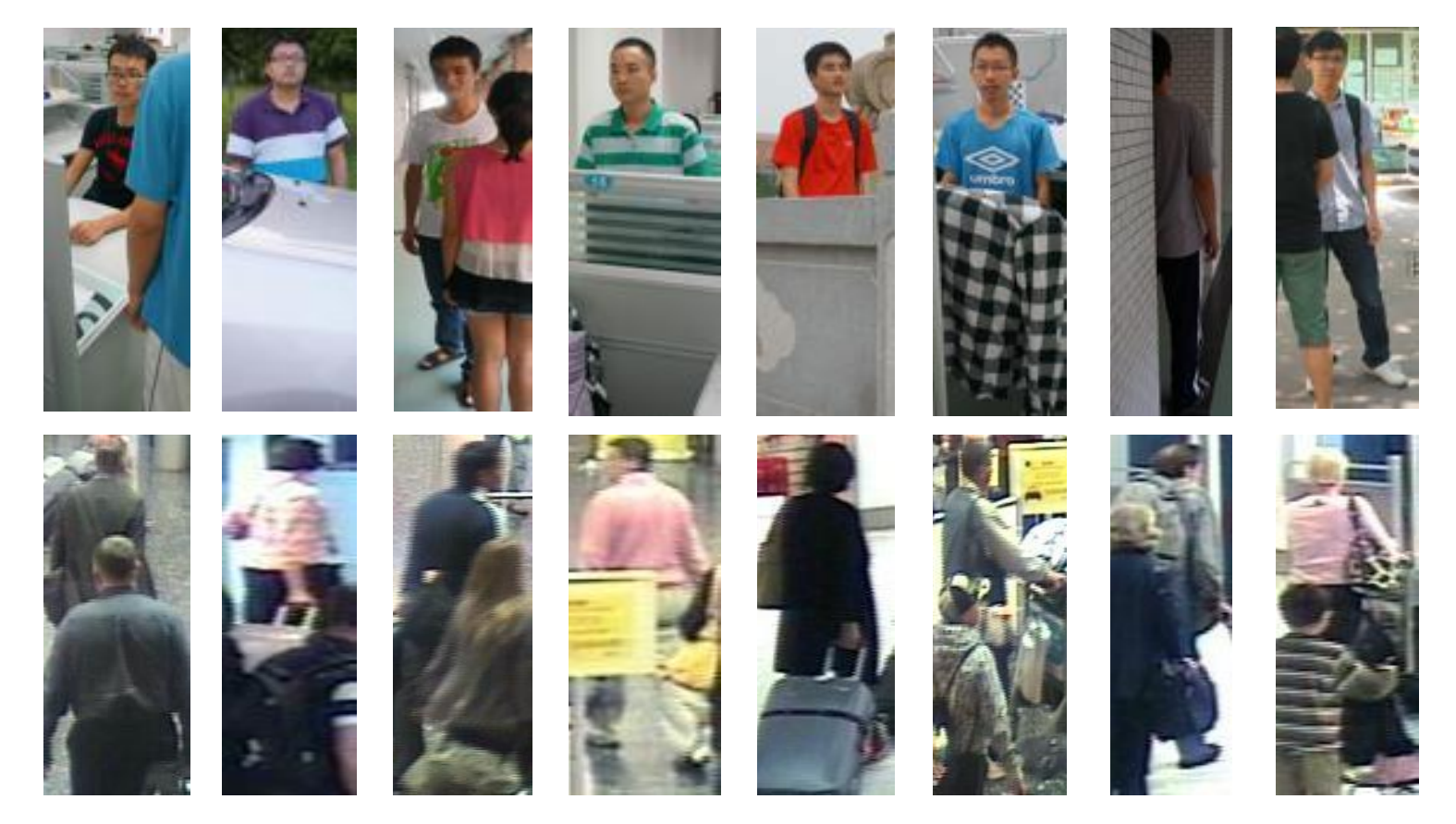}
     \caption{Examples of partial person images.}
    \label{fig:fig1}
    \vspace{-1em}
\end{figure}

Person re-identification (re-id) has witnessed great progress in recent years. Existing approaches generally assume that each image covers a full glance of one person. However, the assumption of person re-id on full and frontal images does not always hold in real-world scenarios, where we merely have access to some partial observations of each person (dubbed partial person images) for idnetification.
For instance, as shown in Fig.~\ref{fig:fig1}, a person in the wild are easily be occluded by moving obstacles (e.g., cars, other persons) and static ones (e.g., trees, barriers), resulting in partial person images.
Hence, partial person re-id has attracted significant research attention as the demand of identification using images captured by CCTV cameras and video surveillance systems has been rapidly growing. However, few studies have focused on identification with partial person images, making partial person re-id an urgent yet unsolved challenging problem.
From this perspective, it is necessary and important for both academic and industrial society to study the partial person re-id problem.

\begin{figure*}[t]
    \centering
       \vspace{0.5em}
    \includegraphics[width=17cm]{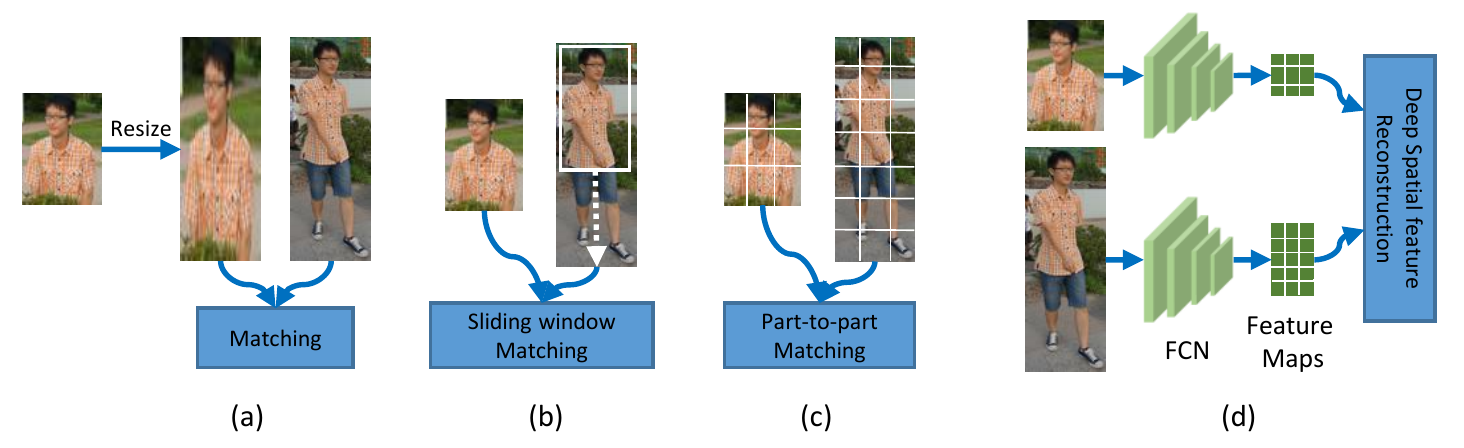}
     \caption{(a) The probe person image and gallery person image are resized to fixed-size (Resizing model). (b) Sliding window matching. (c) Part-based model. (d) The proposed Deep Spatial feature Reconstruction.}
    \label{fig:fig2}
    \vspace{-0.5em}
\end{figure*}

Most existing person re-id approaches fail to identify a person when the body region is severely occluded in the image provided. To match an arbitrary patch of a person, some researchers resort to re-scale an arbitrary patch of the person to a fixed-size image. However, the performance would be significantly degraded due to the undesired deformation (see Fig.~\ref{fig:fig2}(a)). Sliding Window Matching (SWM) \cite{zheng2015partial} indeed introduces a possible solution for partial person re-id by constructing a sliding window of the same size as the probe image and utilizing it to search for the most similar region within each gallery image (see Fig.~\ref{fig:fig2}(b)). However, SWM would not work well when the size of the probe person is bigger than the size of the gallery person. Some person re-id approaches further consider a part-based model which offers an alternative solution of partial person re-id in Fig.~\ref{fig:fig2}(c). However, their computational costs are extensive and they require strict person alignment beforehand. Apart from these limitations, both SWM and part-based models repeatedly extract sub-region features without sharing intermediate computation results, which lead to  unsatisfactory computation efficiency.

In this paper, we propose a novel and fast partial person re-id framework that matches a pair of person images of different sizes (see Fig.~\ref{fig:fig2}(d)). In the proposed framework, Fully Convolutional Network (FCN) is utilized to generate spatial feature maps of certain sized, which can be considered as pixel-level feature matrices. Motivated by the remarkable successes achieved by dictionary learning in face recognition \cite{liao2013partial, wright2009robust,zhang2011sparse}, we develop an end-to-end model named Deep Spatial feature Reconstruction (DSR), which expects that each pixel in the probe spatial maps can be sparsely reconstructed on the basis of spatial maps of gallery images. In this manner, the model is independent of the size of images and naturally avoids the time-consuming alignment step.
Specifically, we design an objective function for FCN which encourages the reconstruction error of the spatial feature maps extracted from the same person to be smaller than that of different identities.
Generally, the major contributions of our work are summarized as follows:
\begin{itemize}
  \item We propose a novel approach named Deep Spatial feature Reconstruction (DSR) for partial person re-id, which is alignment-free and flexible to arbitrary-sized person images.

  \item We first integrate sparse reconstruction learning and deep learning in a unified framework, and train an end-to-end deep model through minimizing the reconstruction error for coupled person images from the same identity and maximizing that of different identities.

  \item Besides, we further replace the pixel-level reconstruction with a block-level one, and develop a multi-scale (different block sizes) fusion model to enhance the performance.

  \item Experimental results demonstrate that the proposed approach achieves impressive results in both accuracy and efficiency on Partial-REID \cite{zheng2015partial} and Partial-iLIDs \cite{zheng2011person} databases.
\end{itemize}

The remainder of this paper is organized as follows. In Sec.~2, we review the related work about FCN, Sparse Representation Classification (SRC), and existing partial person re-id algorithms. Sec.~3 introduces the technical details of deep spatial feature reconstruction. Sec.~4 shows the experimental results and analyzes the performance in computational efficiency and accuracy. Finally, we conclude our work in Sec.~5.

\section{Related Work}
Since the proposed model is a deep feature learning method for partial person re-id based on Fully Convolutional Network and Sparse Representation Classification, we briefly review several related works in this section.

\noindent\textbf{Fully Convolutional Network.} FCN only contains convolutional layers and pooling layers, which has been applied to spatially dense tasks including semantic segmentation \cite{badrinarayanan2015segnet, chen2016deeplab, girshick2014rich, noh2015learning, shelhamer2017fully} and object detection \cite{girshick2015fast, liu2016ssd,  redmon2016you, ren2015faster}. Shelhamer \emph{et al.} \cite{liu2016ssd} introduced a FCN that is trained end-to-end, pixel-to-pixel for semantic segmentation, which outperformed state-of-the-art models without additional machinery. Liu \emph{et al.} \cite{li2016deep} proposed single shot multi-box detector (SSD) based on FCN that can detect objects quickly and accurately. Besides, FCN has also been exploited in visual recognition problems. He \emph{et al.} \cite{he2015spatial} introduced a spatial pyramid pooling (SPP) layer imposed on FCN to produce fixed-length representation from input of arbitrary sizes.

\noindent\textbf{Sparse Representation Classification.} Wright \emph{et al.} \cite{wright2009robust} introduced a well-known method, SRC for face recognition, which achieved promising performance under occlusions and illumination variations. Further studies \cite{gao2010kernel, zhang2011sparse, yang2010gabor, xu2011two} on face recognition with SRC have also been conducted. SRC has been also applied to signal classification \cite{huang2007sparse}, visual tracking \cite{mei2011robust}, and visual classification \cite{yuan2012visual}, etc.

\noindent\textbf{Partial Person Re-identification.} Partial person re-id has become an emerging problem in video surveillance.
Little research has be done to search for a solution for matching arbitrary-sized images presenting only part of the human body. To address this problem, many methods \cite{donahue2014decaf, girshick2014rich} warp an arbitrary patch of an image to a fixed-size image, and then extract fixed-length feature vectors for matching. However, such method would result in undesired deformation. Part-based models are considered as a solution to partial person re-id. Patch-to-patch matching strategy is employed to handle occlusions and cases where the target is partially out of the camera's view. Zheng $\emph{et al.}$ \cite{zheng2015partial} proposed a local patch-level matching model called Ambiguity-sensitive Matching Classifier (AMC) based on dictionary learning with explicit patch ambiguity modeling, and introduced a global part-based matching model called Sliding Window Matching (SWM) that can provide complementary spatial layout information. However, the computation cost of AMC+SWM is rather extensive as features are calculated repeatedly without further acceleration. Furthermore, similar occlusion problems also occur in partial face recognition. Liao \emph{et al.} \cite{liao2013partial} proposed an alignment-free approach called multiple keypoints descriptor SRC (MKD-SRC), where multiple affine invariant keypoints are extracted for facial features representation and sparse representation based classification (SRC) \cite{wright2009robust} is then used for recognition. However, the performance of keypoint-based methods is not quite satisfying with hand-crafted local descriptors. To this end, we propose a fast and accurate method, Deep Spatial feature Reconstruction (DSR), to address the partial person re-id problem.

\begin{figure}[t]
    \centering
       \vspace{0em}
    \includegraphics[width=8cm]{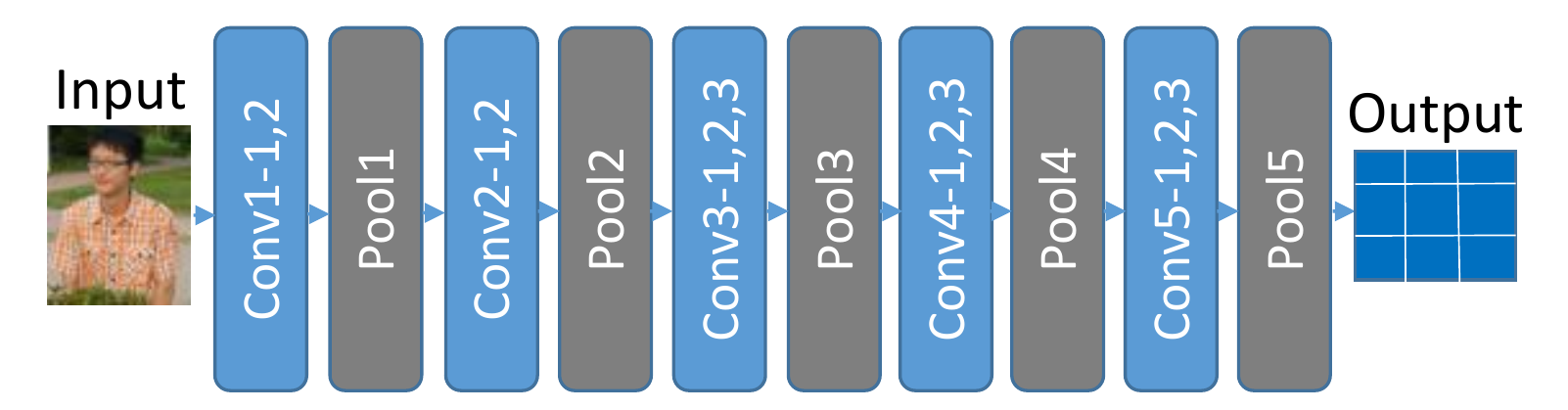}
     \caption{Fully convolutional network.}
    \label{fig:fig3}
    \vspace{0.5em}
\end{figure}

\section{The Proposed Approach}
\subsection{Fully Convolutional Network}
Deep Convolutional Neural Networks (CNNs), as feature extractors in visual recognition task, require a fixed-size input image. However, it is impossible to meet the requirement since partial person images have arbitrary sizes/scales. In fact, the requirement comes from fully-connected layers that demand fixed-length vectors as inputs. Convolutional layers operate in a sliding-window manner and generate correspondingly-size spatial outputs. To handle an arbitrary patch of a person image, we discard all fully-connected layers to implement Fully Convolutional Network that only convolution and pooling layers remain. Therefore, FCN still retains spatial coordinate information, which is able to extract spatial feature maps from arbitrary-size inputs. The proposed FCN is shown in Fig. \ref{fig:fig3}, it contains 13 convolution layers and 5 pooling layers, and the last pooling layer produces identity feature maps.

\subsection{Deep Spatial Feature Reconstruction}
\label{dsfr}
\begin{figure}[t]
    \centering
       \vspace{0em}
    \includegraphics[width=8.2cm]{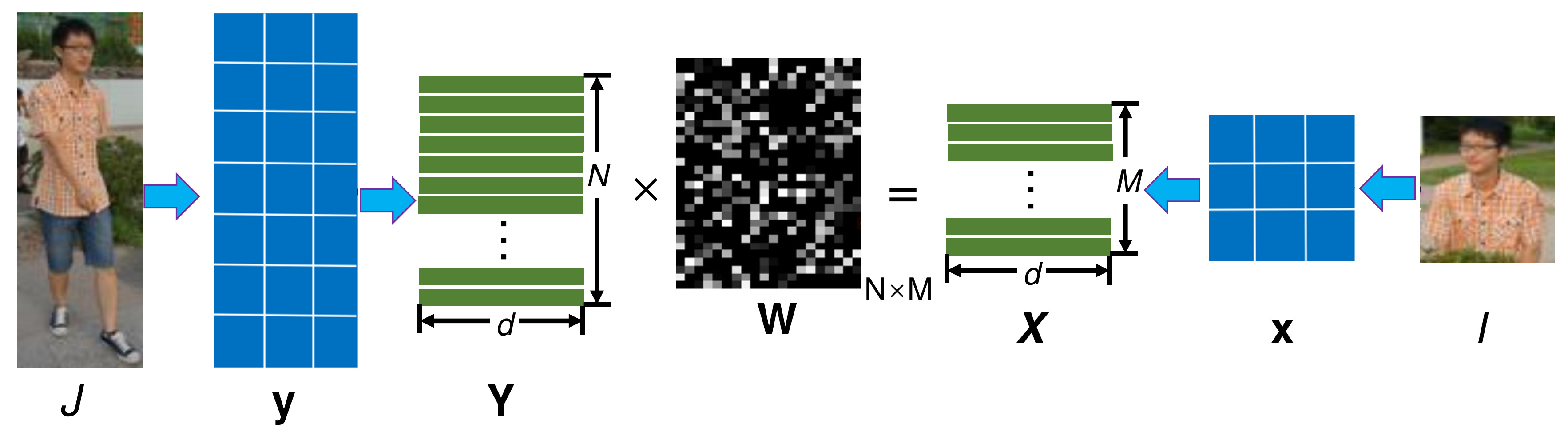}
     \caption{Deep Spatial feature Reconstruction.}
    \label{fig:fig4}
    \vspace{0.5em}
\end{figure}

In this section, we will introduce how to measure the similarity between a pair of person images of different sizes. Assume that we are given a pair of person images, one is an arbitrary patch of person image $I$ (a partial person), and the other is a holistic person image $J$. Correspondingly-size spatial feature maps $\mathbf{x} = conv(I, \theta)$ and $\mathbf{y}=conv(J, \theta)$ are then extracted by FCN, where $\theta$ denotes the parameters in FCN. $\mathbf{x}$ denotes a vectorized $w\times h \times d$ tensor, where $w, h$ and $d$ denote the height, the width and the number of channel of $\mathbf{x}$, respectively. As shown in Fig. \ref{fig:fig4}, we divide $\mathbf{x}$ into $N$ blocks $\mathbf{x}_{n}$, $n=1, \ldots, N$, where $N=w\times h$, and the size of each block is $1\times 1\times d$. Denote by
\begin{equation}
\begin{array}{l}
 \displaystyle \mathbf{X}=\{\mathbf{x}_1, \cdots, \mathbf{x}_N\} \in \mathbb{R}^{d\times N}
\end{array}
\end{equation}
the block set, where $\mathbf{x}_n\in \mathbb{R}^{d\times 1}$. Likewise, $\mathbf{y}$ is divided into $M$ blocks as
\begin{equation}
\begin{array}{l}
 \displaystyle \mathbf{Y}=\{\mathbf{y}_1, \cdots, \mathbf{y}_M\}\in \mathbb{R}^{d\times M},
\end{array}
\end{equation}
then $\mathbf{x}_{n}$ can be represented by linear combination of $\mathbf{Y}$. That is to say, we attempt to search similar blocks to reconstruct $\mathbf{x}_{n}$. Therefore, we wish to solve for the sparse coefficients $\mathbf{w}_n$ of $\mathbf{x}_n$ with respect to $\mathbf{Y}$, where $\mathbf{w}_n \in \mathbb{R}^{M\times 1}$. Since few blocks of $\mathbf{Y}$ are expected for reconstructing $\mathbf{x}_n$, we constrain $\mathbf{w}_n$ using $\ell_1$-norm. Then, the sparse representation formulation is defined as

\begin{equation}
\begin{array}{l}
 \displaystyle \min_{\mathbf{w}_n}||\mathbf{x}_n-\mathbf{Y}\mathbf{w}_n||_2^{2}+\beta||\mathbf{w}_n||_1,
\end{array}
\label{eq3}
\end{equation}

where $\beta$ ($\beta=0.4$ is fixed in our experiment) controls the sparsity of coding vector $\mathbf{w}_n$. $||\mathbf{x}_n-\mathbf{Y}\mathbf{w}_n||_2$ is used to measure the similarity between $\mathbf{x}_n$ and $\mathbf{Y}$. For $N$ blocks in $\mathbf{X}$, the matching distance can be defined as
\begin{equation}
\begin{array}{l}
 \displaystyle d = \frac{1}{N}||\mathbf{X}-\mathbf{Y}\mathbf{W}||_F^{2},
\end{array}
\label{eq4}
\end{equation}
where $\mathbf{W} = \{\mathbf{w}_1, \cdots, \mathbf{w}_N \}\in \mathbb{R}^{M\times N}$ is the sparse reconstruction coefficient matrix. The whole matching procedure is exactly our proposed Deep Spatial feature Matching (DSR). As such, DSR can be used to classify a probe partial person, which does not need additional person alignment. The flowchart of our DSR approach is shown in Fig. \ref{fig:fig4} and the overall DSR approach is outlined in Algorithm \ref{alg:Framwork}.
\begin{algorithm}[t]
\caption{Deep Spatial feature Reconstruction.}
\label{alg:Framwork}
\begin{algorithmic}[1] 
\REQUIRE
A probe person image $I$ of an arbitrary-size; a gallery person image $J$.

\ENSURE Similarity score $d$. \\ 
\STATE Extract probe feature maps $\mathbf{x}$ and gallery feature maps $\mathbf{y}$.
\STATE Divide $\mathbf{x}$ and $\mathbf{y}$ into multiple blocks: $\mathbf{X}=\{\mathbf{x}_1, \cdots, \mathbf{x}_N\}$ and $\mathbf{Y}=\{\mathbf{y}_1, \cdots, \mathbf{x}_M\}$.
\STATE Solve equation (\ref{eq3}) to obtain  sparse reconstruction coefficient matrix $\mathbf{W} = \{\mathbf{w}_1, \cdots, \mathbf{w}_N\}$.
\STATE Solve equation (\ref{eq4}) to obtain similarity score.
\end{algorithmic}
\end{algorithm}

\subsection{Fine-tuning on Pre-trained FCN with DSR}
\label{dsr}
\begin{figure}[t]
    \centering
       \vspace{0em}
    \includegraphics[width=8.2cm]{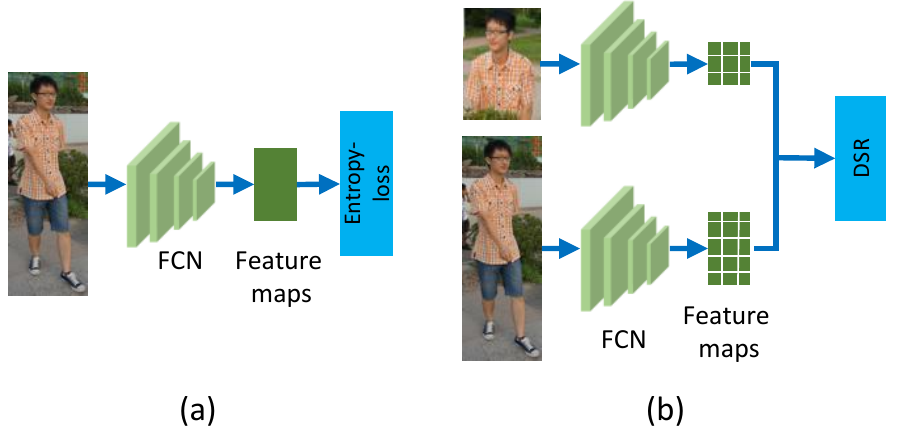}
     \caption{(a) Train FCN with identification signal (entropy-loss). (b) Fine-tune on pre-trained FCN using DSR.}
    \vspace{0.5em}
    \label{fig:fig5}
\end{figure}

 We train the FCN with a particular identification signal that classifies each person images ($320\times 120$ in our experiment) into different identities. Concretely, the identification is achieved by the last pooling layer connected with an entropy-loss (see Fig. \ref{fig:fig5}(a)). To further increase the discriminative ability of deep features extracted by FCN, fine-tuning with DSR is adopted to update the convolutional layers, the framework is shown in Fig. \ref{fig:fig5}(b).

The DSR signal encourages the feature maps of the same identity to be similar while feature maps of the different identities stay away. The DSR can be regarded as verification signal, the loss function is thus defined as
\begin{equation}
\begin{array}{l}
 \displaystyle \mathcal{L}_{veri}(\theta, \mathbf{W})= \alpha||\mathbf{X}-\mathbf{Y}\mathbf{W}||_F^{2}+\beta||\mathbf{W}||_1
\end{array}
\end{equation}
where $\alpha=1$ means that the two features are from the same identity and $\alpha=-1$ for different identities.

We employ an alternating optimization algorithm to optimize $\mathbf{W}$ and $\theta$ in the objective $\mathcal{L}_{veri}$.

\noindent\textbf{Step 1: fix $\mathbf{\theta}$, optimize $\mathbf{W}$}. The aim of this step is to solve sparse reconstruction coefficient matrix $\mathbf{W}$. For solving optimal $\mathbf{W}$, we solve $\mathbf{w}_1, \ldots, \mathbf{w}_N$ respectively, hence, equation (3) is further rewritten as
\begin{equation}
\begin{array}{l}
 \displaystyle \min_{\mathbf{w}_n}\frac{1}{2}\mathbf{w}_n^{T}\mathbf{Y}^{T}\mathbf{Y}\mathbf{w}_n-\mathbf{x}_n^{T}\mathbf{Y}\mathbf{w}_n+\beta||\mathbf{w}_n||_1.
\end{array}
\label{eq:6}
\end{equation}
We utilize the feature-sign search algorithm adopted in \cite{lee2007efficient} to solve an optimal $\mathbf{w}_n$.

\noindent\textbf{Step 2: fix $\mathbf{w}_c$, optimize $\mathbf{\theta}$}. To update the parameters in FCN, we then calculate the gradients of $\mathcal{L}_{veri}(\mathbf{\theta})$ with respect to $\mathbf{X}$ and $\mathbf{Y}$
\begin{equation}
\left\{
\begin{array}{lr}\vspace{0.5em}
\frac{\partial\mathcal{L}_{veri}(\mathbf{\theta})}{\partial{\mathbf{X}}} = 2\alpha (\mathbf{X}-\mathbf{Y}\mathbf{W})\\
\frac{\partial\mathcal{L}_{veri}(\mathbf{\theta})}{\partial{\mathbf{Y}}}= - 2\alpha(\mathbf{X}-\mathbf{Y}\mathbf{W})\mathbf{W}^{T}.
\end{array}
\right.
\vspace{0.5em}
\end{equation}
Clearly, FCN supervised by DSR is trainable and can be optimized by standard Stochastic Gradient Descent (SGD). In Algorithm \ref{alg:Framwork1}, we summarize the algorithm details of feature learning with DSR.

\begin{algorithm}[t]
\caption{Feature Learning with DSR.}
\label{alg:Framwork1}
\begin{algorithmic}[1] 
\REQUIRE Training data $I$ and $J$. The parameter of indicator value $\alpha$ and sparsity strength $\beta$. Pre-trained FCN parameter $\theta$.

\ENSURE FCN parameter $\theta$. \\ 
\STATE Extract multiple blocks $\mathbf{X}$ and $\mathbf{Y}$.
\STATE $t+1 \leftarrow t$
\STATE Compute the reconstruction error by $\mathcal{L}_{veri}(\mathbf{W}, \mathbf{\theta})$.
\STATE Update the  sparse reconstruction coefficient matrix $\mathbf{W}$ using Equation (\ref{eq:6}).
\STATE Update the gradients of $\mathcal{L}_{veri}(\mathbf{W}, \mathbf{\theta})$ with respect to $\mathbf{X}$ and $\mathbf{Y}$.
\STATE Update the parameters $\mathbf{\theta}$ by $\mathbf{\theta}^{t+1}=\mathbf{\theta}^{t}-\alpha(\frac{\partial{\mathcal{L}_{veri}}}{\partial{\mathbf{X}}}\frac{\partial{\mathbf{X}}}{\partial{\mathbf{\theta}^{t}}}+
\frac{\partial{\mathcal{L}_{veri}}}{\partial{\mathbf{Y}}}\frac{\partial{\mathbf{Y}}}{\partial{\mathbf{\theta}^{t}}})$
\STATE \textbf{end while}
\end{algorithmic}
\end{algorithm}

We directly embed the proposed DSR into FCN to train an end-to-end deep network, which can improve the overall performance. It is noteworthy that person images in each training pair share the same scale.

\subsection{Multi-scale Block Representation}
\begin{figure}[t]
    \centering
       \vspace{0.0em}
    \includegraphics[width=8.2cm]{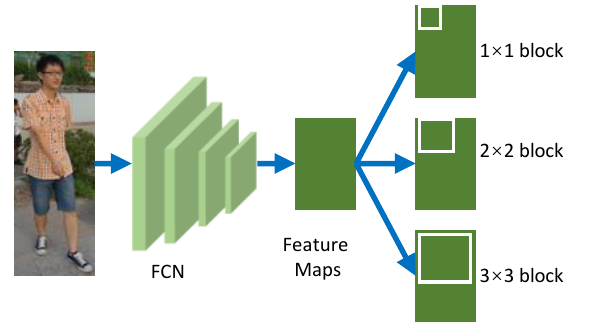}
     \caption{Multi-scale block representation.}
    \vspace{-0.5em}
    \label{fig:fig6}
\end{figure}

Extracting features that are invariant to probe images with unconstrained scales are challenging and important for solving partial person re-id problem.
Unlike holistic person images where we can directly resize the image are of the whole person to a fixed size, it is difficult to determine the scale of the person occluded in probe image explicitly. Therefore, the scales between a partial person and a holistic person are vulnerable to mismatching, which would result in performance degradation. Single-scale blocks (1$\times$1 blocks) used in Sec.~\ref{dsfr} are not robust to scale variations. To alleviate the influence of scale mismatching, multi-scale block representation is also proposed in DSR (see Fig. \ref{fig:fig6}). In our experiments, we adopt blocks of 3 different scales: 1$\times$1, 2$\times$2 and 3$\times$3, and these blocks are extracted in a sliding-window manner (stride is 1 block).

In order to keep the dimensions consistent, 2$\times$2 and 3$\times$3 blocks are resized to 1$\times$1 block by average pooling. The resulting blocks are all pooled in the block set. The main purpose of multi-scale block representation is to improve the robustness against scale variation. Experiment results show that such processing operations can effectively improve the performance the proposed method.

Unlike some region-based models that perform multi-scale operations in image-level, suffering from expensive computation cost due to calculating features repeatedly, the proposed multi-scale block representation is conducted in feature-level, which greatly reduce the computational complexity as the features are only computed once and shared among different block dividing patterns.

\section{Experiments}
In this section we mainly focus on seven aspects below, 1). exploring the influence of deformable person images; 2).  the benefits of multi-scale block representation; 3). comparisons with other partial person re-id approaches; 4). computational time of various partial person re-id approaches; 5). effectiveness of fine-tuning with DSR. 6). evaluation on holistic person image.

\subsection{Experiment Settings}
\noindent\textbf{Network Architecture.} The designed Fully Convolutional Network (FCN) is shown in Fig. \ref{fig:fig3}. The Market1501 dataset \cite{wang2012query} is used to pre-train the FCN followed by a 1,500-way softmax with the size of network input set to $320\times120$. 3,000 positive pairs of person images and 3,000 negative pairs of person images are used to fine-tune the pre-trained FCN with DSR. For each pair, one is a holistic person image and the other is an arbitrary patch of a person image.

\begin{figure}[t]
    \centering
       \vspace{0em}
    \includegraphics[width=8.4cm]{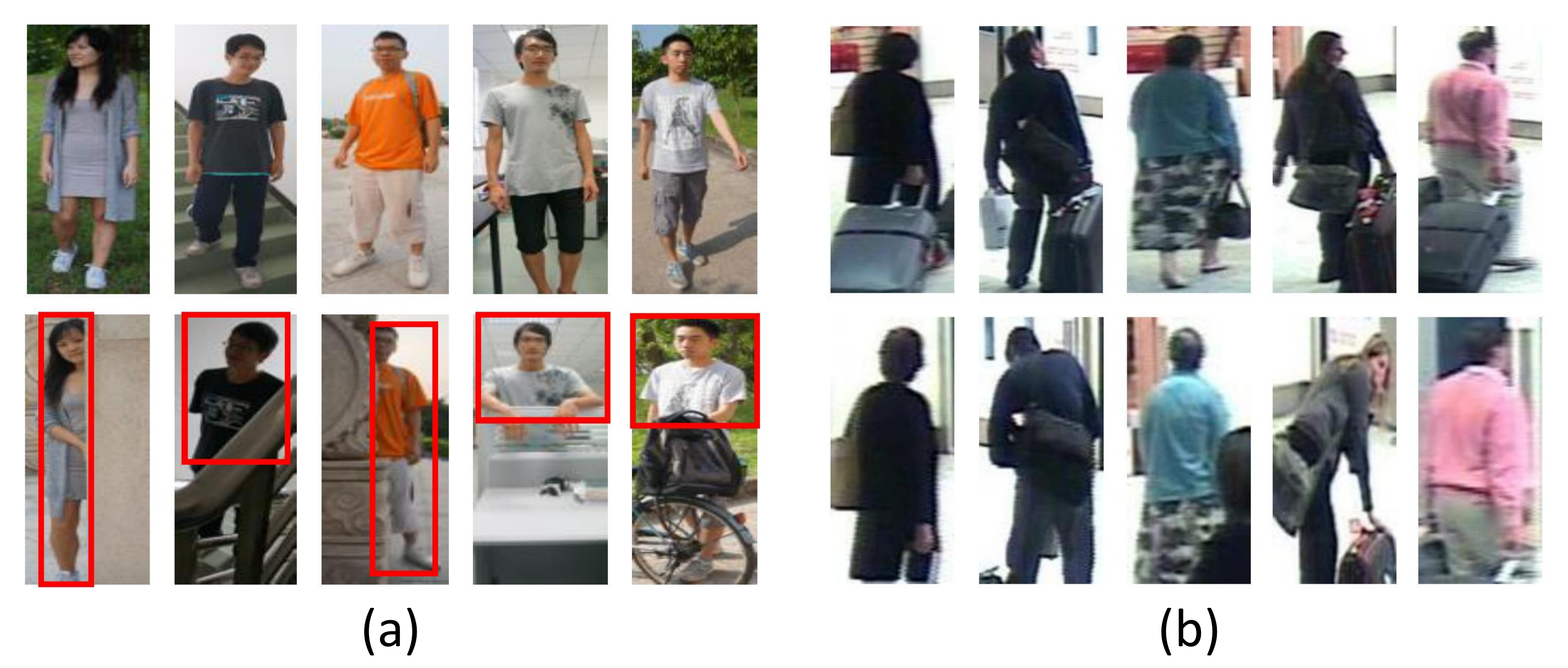}
     \caption{Examples of partial persons in Partial REID (a) and P-iLIDS Dataset (b) Datasets.}
    \vspace{-0.5em}
    \label{fig:fig7}

\end{figure}

\noindent\textbf{Datasets.} Partial REID dataset is a specially designed partial person dataset that includes 600 images from 60 people, with 5 full-body images and 5 partial images per person. These images are collected at a university campus from different viewpoints, backgrounds and different types of severe occlusion. The examples of partial persons in the Partial REID dataset are shown in Fig. \ref{fig:fig7}(a). The region in the red bounding box is the partial person image. The probe set consists of all partial images per person, and the holistic person images are used as the gallery set. Partial-iLIDS is a simulated partial person dataset based on iLIDS \cite{zheng2011person}. The iLIDS contains a total of 476 images of 119 people captured by multiple non-overlapping cameras. Some images in the dataset contain people occluded by other individuals or luggages. Fig. \ref{fig:fig7}(b) shows some examples of individual images from the iLIDS dataset. For the occluded individuals, the partial observation is generated by cropping the non-occluded region of one image of each person to construct the probe set. The non-occluded images of each person are selected to construct a gallery set. There are $p=60$ and $p=119$ individuals in each test set for the Partial REID and Partial-iLIDS datasets respectively. One and five partial person images of each person are used as a probe set for the Partial REID and Partial-iLIDS datasets, respectively.

\renewcommand{\arraystretch}{1.1}
\begin{table}[t]
\small
\centering
  \label{table7}
    \begin{tabular}{|c|c|c|c|c|}
    \hline
    Dataset&Individual &Image & Gallery& Probe\cr
    \hline

    Partial REID \cite{zheng2015partial}&60&600 &5& 5\cr \hline
    Partial-iLIDS \cite{zheng2011person}&119&476&3& 1\cr
    \hline
    \end{tabular}

\end{table}

\noindent\textbf{Evaluation Protocol.} In order to show the performance of the proposed approach, we provide the average Cumulative Match Characteristic (CMC) curves for close-set experiments and Receiver Operating Characteristic (ROC) curves for verification experiments to evaluate our algorithm.

\noindent\textbf{Benchmark Algorithms.} Several existing partial person re-id methods are used for comparison, including part-based matching method Ambiguity-sensitive Matching (AMC) \cite{zheng2015partial}, global-to-local matching method Sliding Window Matching (SWM) \cite{zheng2015partial}, AMC+SWM \cite{zheng2015partial} and Resizing model (see Fig. \ref{fig:fig2}(a)). For AMC, features are extracted from $64\times 64$ supporting areas, and these supporting areas are densely sampled with an overlap of half of the height/width of the supporting area in both horizontal and vertical directions. Each region is represented by the fine-tuning FCN, creating a 2,048-dimension feature vector (the output size is $2\times2\times512$ for the proposed FCN).

\noindent\textbf{Settings.} Both single-shot and multi-shot experiments are conducted. $N$ person images ($N=1$ for single-shot experiments and $N>1$ for multi-shot experiments) are used as gallery images for each individual.

\subsection{Influence of Person Image Deformation}
Fig. \ref{fig:fig2}(a) shows the details of the Resizing model, where person images in the gallery and probe set are all re-sized to $320\times120$. FCN is used as the feature extractor and 15,360-dimension feature vector is produced for each person image. In the single-shot experiments, we use Euclidean distance to measure the similarity of a pair of person images in the Resizing model. In the multi-shot experiments, we return the average similarity between the probe person image and multiple gallery images of an certain individual. For DSR, we only adopt single-scale block representation ($1\times1$ block) in this experiment.
Table \ref{table7} shows the experimental results on Partial REID and Partial-iLIDS datasets. 
It is clear that DSR consistently outperfoms the Resizing model across all experiment settings. Such results justifies the fact that person image deformation would significantly affect the recognition performance.
For example, resizing the upper part of a person image to a fixed-size would cause the the entire image to be stretched and deformed.
\renewcommand{\arraystretch}{1.1}
\begin{table}[t]
  \footnotesize
  \centering
  \caption{Influence of person image deformation (rank-1 accuracy).}
  \label{table7}
    \begin{tabular}{|l|c|c|c|c|}
    \hline
    \multirow{2}{*}{Method}&
    \multicolumn{2}{c|}{Partial REID}&\multicolumn{2}{c|}{Partial-iLIDS}\cr\cline{2-5}
    &$N=1$&$N=3$&$N=1$ &$N=3$ \cr
    \hline
    Resizing model &19.33&26.00&21.85&28.57 \cr
    DSR &\bf 39.33&\bf 49.33&\bf 51.06 &\bf 54.58\cr\hline
    \end{tabular}
\end{table}

\begin{figure}[t]
    \centering
       \vspace{0em}
    \includegraphics[width=8.55cm]{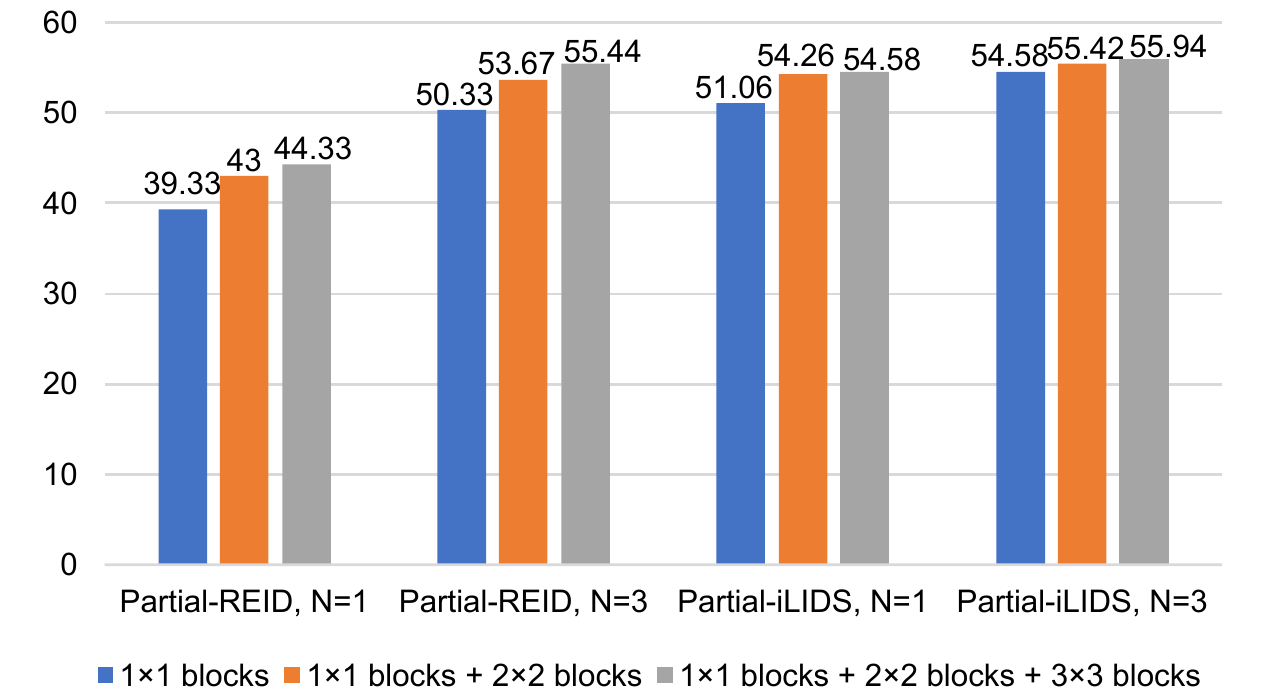}
     \caption{Rank-1 accuracy of DSR with single-scale block representation and multi-scale block representation.}
    \label{fig:fig9}
    \vspace{-0.5em}
\end{figure}

\subsection{Multi-scale Block Representation Benefits}
To evaluate the performance of the proposed DSR with regard to the multi-scale block representation, we pool different-size blocks into the gallery and probe block set. 3 different fusion ways are adopted: $1\times1$ blocks, $1\times1$ blocks combined with $2\times2$ and $1\times1$ blocks, $2\times2$ blocks combined with $3\times3$ blocks. Results are shown in Fig. \ref{fig:fig9}. DSR achieve the best performance when gallery and probe block set contain $1\times1$, $2\times2$ and $3\times3$ blocks. Experimental results suggest that multi-scale block representation is effective. The single-scale block contains more local information, while the multi-scale block is able to provide complementary information to make DSR more robust to scale variation.

\subsection{Comparison to the State-of-the-Art}
We compare the proposed DSR to the state-of-the-art methods, including AMC, SWM, AMC+SWM and Resizing model, on the Partial REID and Partial-iLIDS datasets. There are $p=60$ and $p=119$ individuals in each of the test sets for the Partial REID and Partial-iLIDS datasets respectively. For DSR, we report the results using single-scale block representation and multi-scale bloc representation. For AMC+SWM, the weights of AMC and SWM are 0.7 and 0.3, respectively. Both the single-shot setting and the multi-shot setting are conducted in this experiment.

\begin{table}[t]
  \small
  \centering
  \caption{Performance comparison on single-shot experiment.}
  \label{table6}
    \begin{tabular}{|l|c|c|c|c|c|c|}
    \hline
    \multirow{2}{*}{Method}&
    \multicolumn{2}{c|}{Partial REID}&\multicolumn{2}{c|}{Partial-iLIDS}\cr\cline{2-5}
    &$r=1$&$r=3$ &$r=1$&$r=3$ \cr
    \hline
    Resizing model &19.33&32.67&21.85&36.97 \cr
    SWM \cite{zheng2015partial}&24.33&45.00&33.61&47.06\cr
    AMC \cite{zheng2015partial}&33.33&46.00&46.78&64.75 \cr
    AMC+SWM \cite{zheng2015partial}&36.00&51.00&49.58&63.34\cr \hline
    DSR (single-scale)& 39.33& 55.67 &51.06& 61.66\cr
    DSR (multi-scale)&\bf 43.00&\bf 60.33&\bf 54.58&\bf 64.50\cr\hline
    \end{tabular}
\end{table}

\begin{figure}[t]
    \centering
       \vspace{-0.5em}
    \includegraphics[width=7cm]{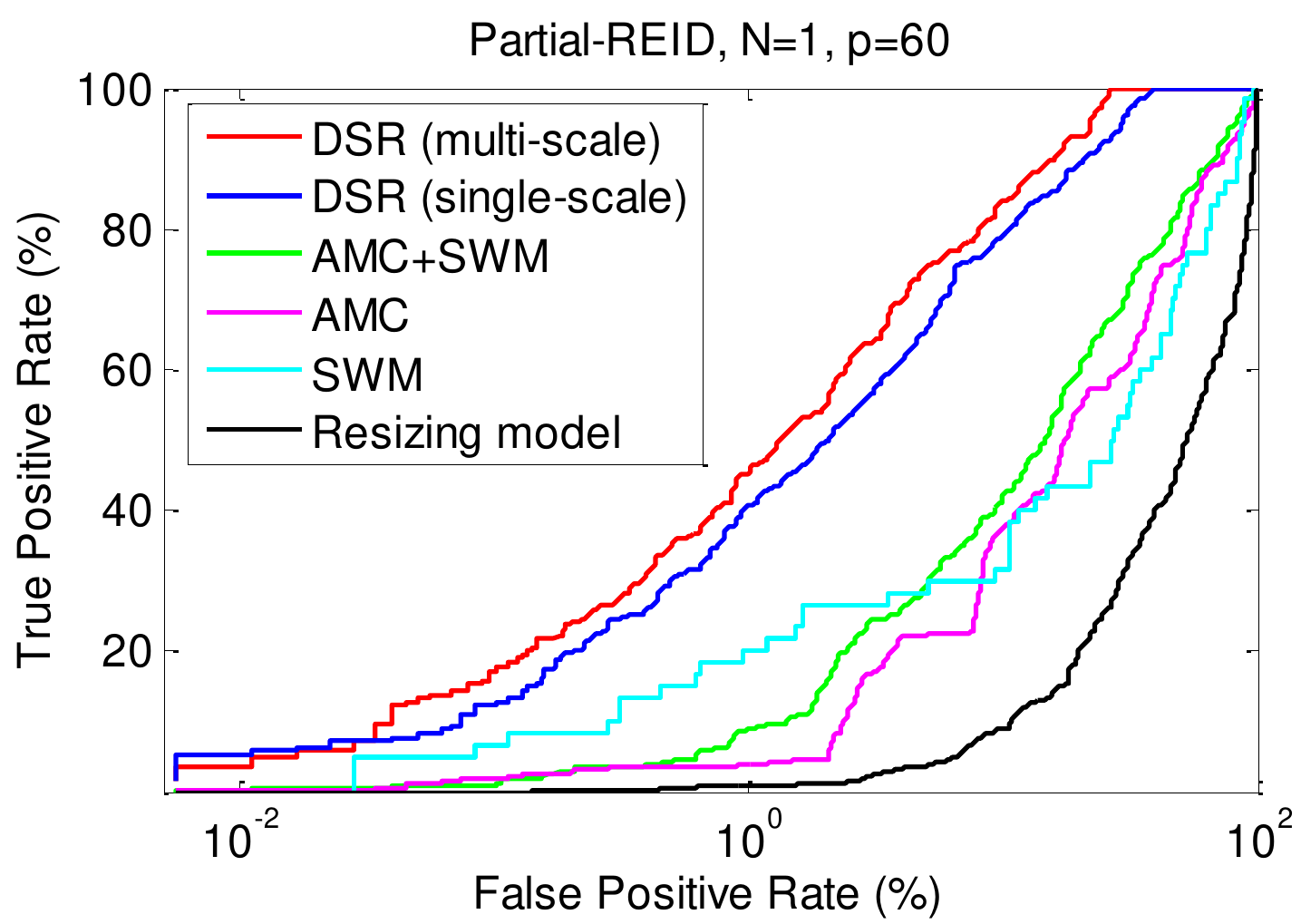}
     \caption{ROC curves of various partial person re-id approaches on Partial REID Dataset.}
    \vspace{0em}
    \label{fig:fig11}
 \vspace{-0.6em}
\end{figure}

\begin{figure}[t]
    \centering
       \vspace{-0.3em}
    \includegraphics[width=7cm]{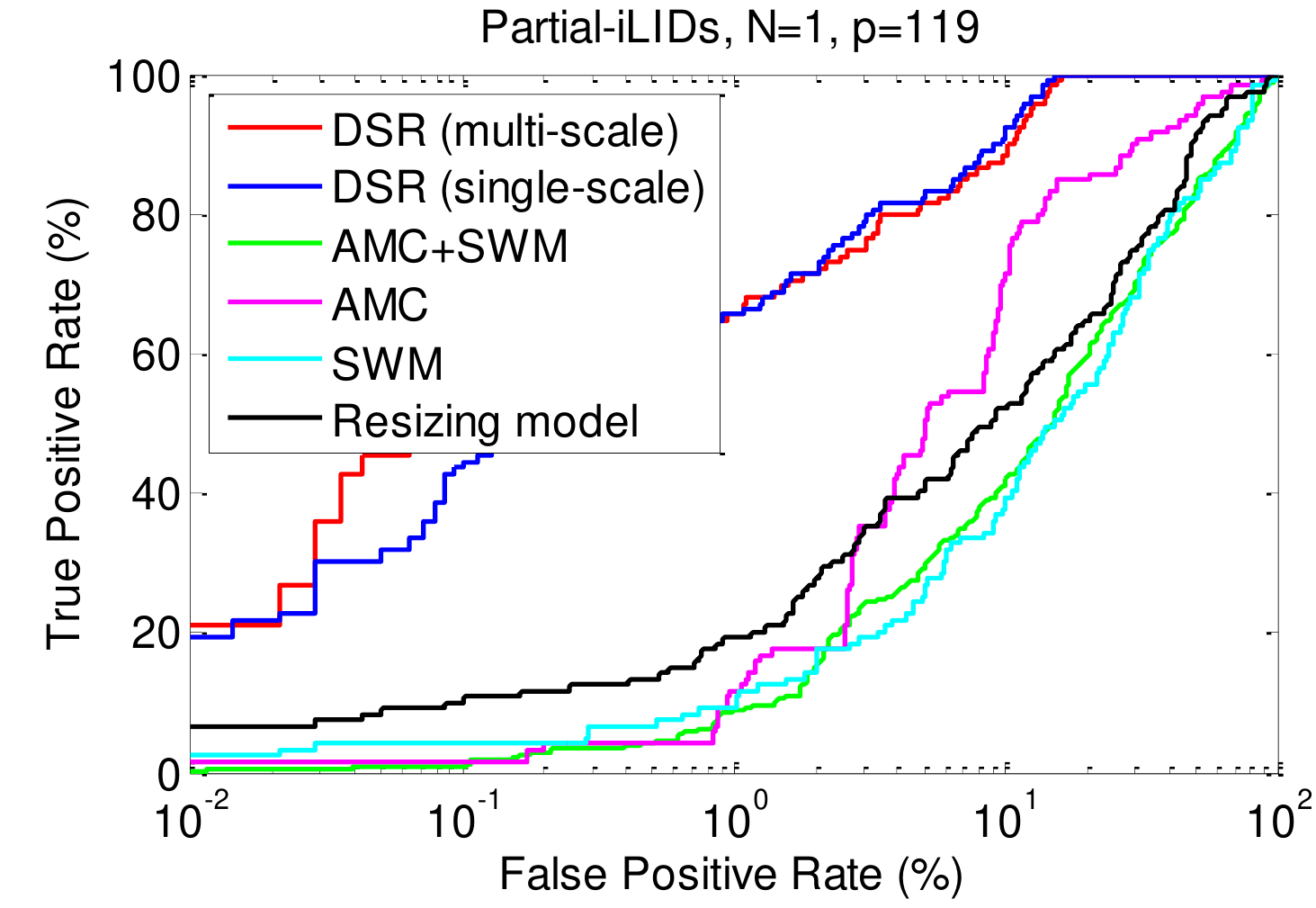}
     \caption{ROC curves of various partial person re-id approaches on Partial-iLIDS Dataset.}
    \vspace{0em}
    \label{fig:fig12}
 \vspace{-1em}
\end{figure}

\noindent\textbf{Single-shot experiments.} Table \ref{table6} shows the single-shot experimental results. We find the results on Partial REID and Partial-iLIDS are similar. The proposed method DSR outperforms AMC, SWM, AMC+SWM and Resizing model. DSR takes full advantage of FCN that operate in a sliding-window manner and outputs feature maps without deformation. AMC is a local-to-local matching method that achieves comparable performance because background patches can be automatically excluded due to their low visual similarity. Thus, it is somewhat robust to occlusion. However, it is difficult to select satisfactory support area size and stride making it not robust to scale variation. SWM is a local-to-global matching method, which requires that the probe size is smaller than the gallery size. Search manner in SWM would ignore some detailed information about a person image. AMC+SWM perform as well as DSR because local features in AMC combined with global features in SWM makes it robust to occlusion and view/pose various. Similar results are also observed from the ROC curves shown in Fig.~\ref{fig:fig11} and Fig.~\ref{fig:fig12}. Obviously, DSR shows small intra-distance and large inter-distance.

\begin{figure}[t]
    \centering
       \vspace{0em}
    \includegraphics[width=8.4cm]{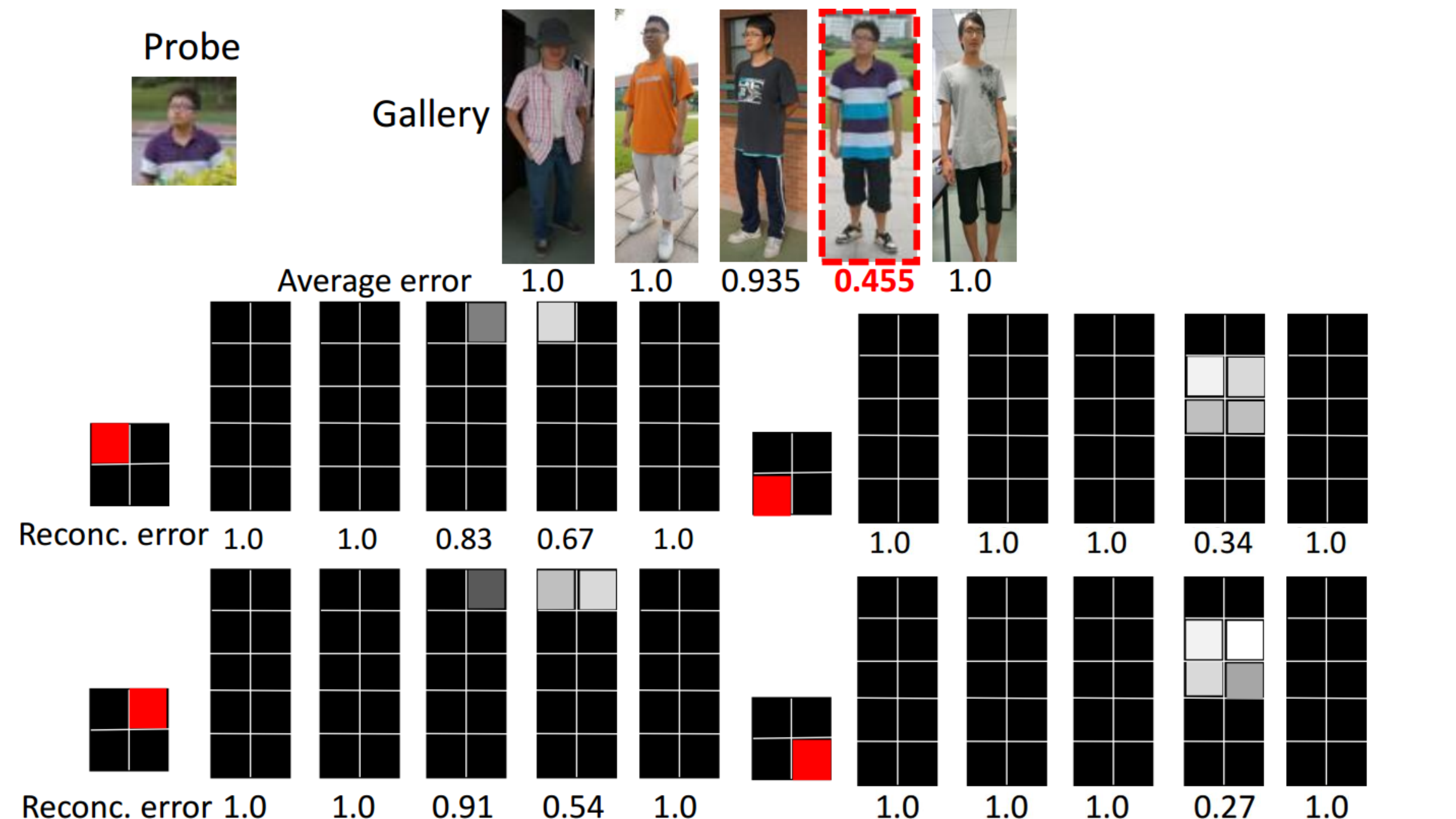}
     \caption{Examples of searching similar blocks.}
    \vspace{-0.7em}
    \label{fig:fig10}
\end{figure}

\begin{table}[t]
  \small
  \centering
  \caption{Performance comparison on multi-shot experiment.}

    \begin{tabular}{|l|c|c|c|c|}
    \hline
    \multirow{2}{*}{Method}&
    \multicolumn{2}{c|}{Partial REID}&\multicolumn{2}{c|}{Partial-iLIDS}\cr\cline{2-5}
    &$r=1$&$r=3$&$r=1$&$r=3$ \cr
    \hline
    Resizing model &26.00&37.00&28.57&43.67 \cr
    SWM \cite{zheng2015partial}&34.33&47.67&35.33&49.67\cr
    AMC \cite{zheng2015partial}&42.33&55.67&44.67&56.33 \cr
    AMC+SWM \cite{zheng2015partial}&44.67&56.33&52.67&63.33\cr\hline
    DSR (single-scale)& 49.33&65.67 &54.67&64.33\cr
    DSR (multi-scale)&\bf 53.67&\bf 72.33 &\bf 55.46&\bf 68.07\cr\hline

    \end{tabular}
    \label{tab6}
\end{table}
As shown in Fig. \ref{fig:fig10}, we illustrate the solution for searching the most similar person image to an input probe image. Four blocks are respectively reconstructed by all blocks from gallery feature maps, then the reconstruction errors are averaged to find the minimum one. Looking carefully the reconstruction coefficients, the feature blocks from the probe could be well reconstructed by similar feature blocks from the gallery image of the same identity. Even though the size of the gallery image or the postion and viewpoint of the person in the gallery image are not consistent with that of the probe image, we could still use DSR to find similar gallery blocks to reconstruct probe blocks, and finally obtain the minimum reconstruction error.

\noindent\textbf{Multi-shot experiments.} DSR approach is evaluated under the multi-shot setting (N=3) on Partial REID and Partial-iLIDS datasets. The results are shown in Table \ref{tab6}. Similar results are obtained in the single-shot experiment. Specifically, the results show that multi-shot setup helps to improve the performance of DSR since DSR increases from 39.33\% to 49.33\% on Partial REID dataset and from 51.06\% to 54.67\% on Partial-iLIDS dataset.

\subsection{Computational Efficiency}
Our implementation is based on the publicly available code of \emph{MatConvnet} \cite{vedaldi2015matconvnet}. All experiments in this paper are trained and tested on PC with 16GB RAM, i7-4770 CPU @ 3.40GHz. Single-shot and multi-shot experiments on Partial REID dataset are conducted to test the computational time of identifying a probe person image. For DSR, we use single-scale block representation ($1\times1$ block) and multi-scale block representation ($1\times1$ and $2\times2$ blocks). Table \ref{table3} shows the computational time of various partial person re-id approaches, which suggests that the propose DSR outperforms other approaches in computation efficiency. DSR with single-scale block representation and multi-scale block representation respectively take 0.269s and 0.278s to identify a person image. For AMC, it costs more computational time than DSR because it repeatedly runs FCN for each sub-region without sharing computation. For SWM, it sets up a sliding window of the same as the probe person image to search for similar sub-region within each gallery image. Generally, many sub-regions would generate by the sliding window, which increases extensive computational time of feature extraction. Besides, when given a new probe person image, it requires regenerating sub-region by the sliding window of the same as the probe image. DSR performs better than the Resizing model, the computational cost of feature extraction would increase after resizing.

\begin{table}[t]
  \small
  \centering
  \caption{Computational time comparison on Partial REID dataset.}
  \vspace{0.5em}
  \label{table3}
    \begin{tabular}{|l|c|c|}
    \hline
    \multirow{2}{*}{Method}&
    \multicolumn{2}{c|}{Computational time (s)}\cr\cline{2-3}
    &$N=1$ &$N=3$ \cr
    \hline
    Resizing model &0.326 & 0.371\cr
    AMC \cite{zheng2015partial}&0.972 & 1.213\cr
    SWM \cite{zheng2015partial}&81.519 & 237.144\cr\hline
    DSR (single-scale)&\bf 0.269 &\bf  0.265\cr
    DSR (multi-scale)&0.278&0.285 \cr
    \hline
    \end{tabular}
    \vspace{0em}
\end{table}

\subsection{Contribution of Fine-tuning with DSR}
\begin{figure}[t]
    \centering
       \vspace{0.0em}
    \includegraphics[width=8.5cm]{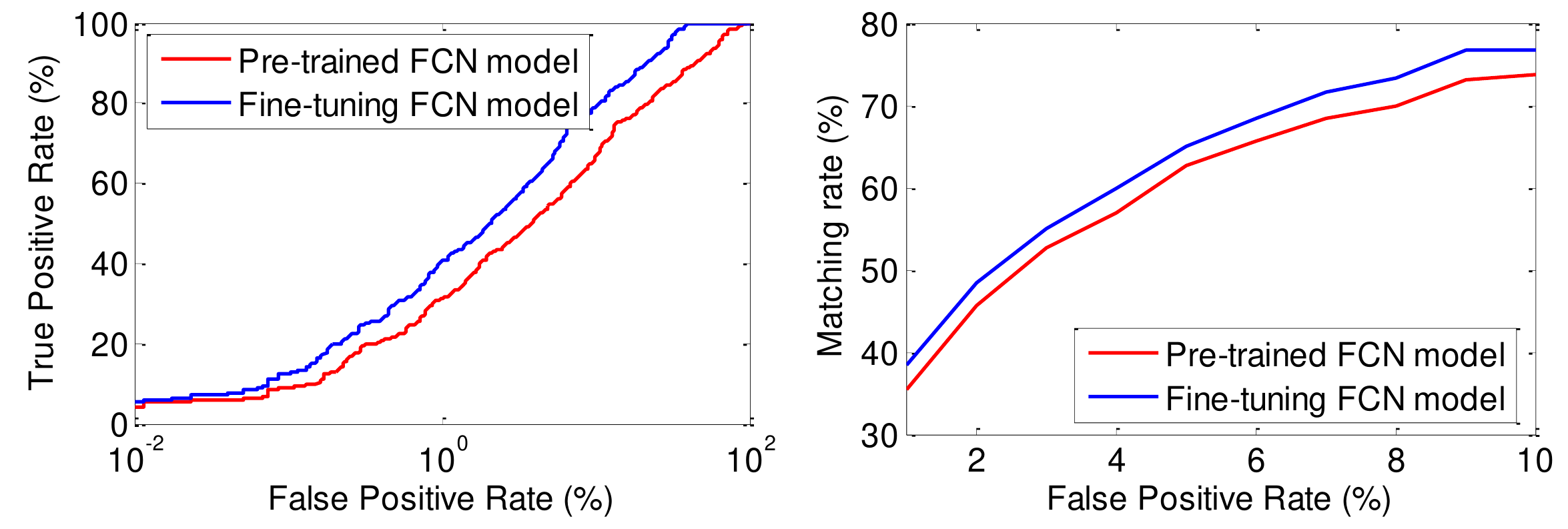}
    \vspace{-1em}
     \caption{ROC curves and CMC curves on Partial REID dataset using pre-trained FCN model and fine-tuning FCN model (N=1).}
    \vspace{-1em}
    \label{fig:fig13}
 \vspace{0em}
\end{figure}

In section \ref{dsr}, DSR is used to fine-tune on the pre-trained FCN to learn more discriminative spatial features. To verify the effectiveness of fine-tuning FCN with DSR, we conduct the single-shot experiment on Partial REID dataset. We compare the pre-trained FCN (FCN training only with softmax loss is regarded as a pre-trained model) to the fine-tuning FCN with DSR (fine-tuning model). Fig. \ref{fig:fig13} shows ROC curves and CMC curves of these two models. Experimental results show that the fine-tuning FCN model performs better than the pre-trained model, which indicates that fine-tuning with DSR can learn more discriminative spatial deep features. Pre-trained model with softmax loss training can only represent the probability of each class that a person image belongs to. For the fine-tuning model, DSR can effectively reduce the intra-variations between a pair of person images of the same individual.

\subsection{Evaluation on Holistic Person Image}
To verify the effectiveness of DSR on holistic person re-identification, we carry out additional holistic person re-id experiments on Market1501 dataset \cite{zheng2015scalable}. Market1501 is one of the largest benchmark dataset that contains 1,501 individuals which are captured by six surveillance cameras in campus. Each individual is captured by two disjoint cameras. Totally it consists of 13,164 person images and each individual has about 4.8 images at each \textbf{viewpoint}. We follow the standard test protocol, i.e., 751 individuals are used for training and 750 individuals are used for testing. The ResNet50 pre-trained on ImageNets is used as the baseline model. For DSR, feature maps extracted from $\emph{res5c}$ are used as identity feature. We respectively adopt single-scale representation ($1\times 1$) and multi-scale representation ($1\times 1$, $2\times 2$ and $3\times3$) in feature representation term. Experimental results in Table \ref{table4} suggest that DSR achieves the best performance. We draw three conclusions: 1) DSR is very effective compared to Euclidean distance because DSR can automatically search similar feature blocks for best matching; 2) multi-scale presentation can achieve better results because it avoids the influence of scale variations; 3) training model with DSR effectively learns more discriminative deep spatial features, which encourages the feature maps of the same identity to be similar while feature maps of the different identities are pushed far apart.
\begin{table}[t]
  \small
  \centering
  \caption{Experimental results on Market1501 with single query.}
  \vspace{0.5em}
  \label{table4}
    \begin{tabular}{|l|c|c|}
    \hline
    Method & $r=1$  & mAP\\
    \hline
    BOW \cite{zheng2015scalable}& 34.38 & 14.10 \cr
    MSCAN \cite{li2017learning}& 80.31 & 57.53 \cr
    Spindle \cite{zhao2017spindle}& 76.90 & - \cr
    Re-ranking \cite{zhong2017re}& 77.11 & 63.63 \cr
    CADL \cite{lin2017consistent}& 80.85 & 55.58 \cr
    CAMEL \cite{yu2017cross} & 54.50 & 26.30 \cr
    DNSL+OL-MANS \cite{zhou2017efficient}& 60.67 & - \cr
    DLPAR \cite{zhao2017deeply}&81.00&-\cr

    \hline
    Resnet50-pool5  & \multirow{2}{*}{77.40} & \multirow{2}{*}{55.64} \cr
    +Euclidean distance (baseline model) & &\cr
    \hline
    Resnet50-res5c (single-scale)+DSR&{82.72} & {61.25}\cr
    \hline

    Resnet50-res5c (multi-scale)+DSR&{\bf 83.58} & {\bf 64.25}\cr
    \hline
    \end{tabular}
    \vspace{0em}
\end{table}

\section{Conclusion}
We have proposed a novel approach called Deep Spatial feature Reconstruction (DSR) to address partial person re-identification. To get rid of the fixed input size, the proposed spatial feature reconstruction method provides a feasibility scheme where each channel in the probe spatial feature map is linearly reconstructed by those channels of a gallery spatial image map, it also avoids the trivial alignment-free matching.
Furthermore, we embed DSR into FCN to learn more discriminative features, such that the reconstruction error for a person image pair from the same person is minimized and that of image pair from different persons is maximized.
Experimental results on the Partial REID and Partial-iLIDS datasets validate the effectiveness and efficiency of DSR, and the advantages over various partial person re-id approaches are significant. Additionally, the proposed method is also competitive in the holistic person dataset, Market1501. \\

\textbf{Acknowledgments}  This work is supported by the Beijing Municipal Science and Technology Commission (Grant No. Z161100000216144) and National Natural Science Foundation of China (Grant No. 61427811, 61573360). Special thanks to Dangwei Li and Yunfan Liu who support our experiments.
{
\bibliographystyle{ieee}
\bibliography{egbib}
}

\end{document}